\documentclass{ifacconf}

\usepackage{graphicx}      % include this line if your document contains figures
\usepackage{natbib}        % required for bibliography

\usepackage{amsmath,amssymb}

\usepackage{algpseudocode}
\usepackage{algorithm}
\algnewcommand{\LeftComment}[1]{\Statex \(\triangleright\) #1}

\usepackage{url}

\begin{document}
\begin{frontmatter}

\title{Blending of Learning-based Tracking and Object Detection for Monocular Camera-based Target Following}

%\title{Style for IFAC Conferences \& Symposia: Use Title Case for
%  Paper Title\thanksref{footnoteinfo}} 

%\thanks[footnoteinfo]{Sponsor and financial support acknowledgment
%goes here. Paper titles should be written in uppercase and lowercase
%letters, not all uppercase.}

\author[First]{Pranoy Panda} 
\author[Second]{Martin Barczyk} 

\address[First]{Department of Electronics and Communication Engineering, National Institute of Technology, Rourkela, Odisha 769008, India (e-mail: 
1997pranoy@gmail.com)}
\address[Second]{Department of Mechanical Engineering, University of Alberta,
        Edmonton, AB, T6G 1H9, Canada (e-mail: mbarczyk@ualberta.ca)}

\begin{abstract}                % Abstract of not more than 250 words.
Deep learning has recently started being applied to visual tracking of generic objects in video streams. For the purposes of robotics applications, it is very 
important for a target tracker to recover its track if it is lost due to heavy or prolonged occlusions or motion blur of the target. We present a real-time 
approach which fuses a generic target tracker and object detection module with a target re-identification module. Our work focuses on improving the performance 
of Convolutional Recurrent Neural Network-based object trackers in cases where the object of interest belongs to the category of \emph{familiar} objects. Our 
proposed approach is sufficiently lightweight to track objects at 85-90 FPS while attaining competitive 
results on challenging benchmarks.
\end{abstract}


\end{frontmatter}
%===============================================================================

\section{Introduction}
\label{sec:intro}

Object tracking is an essential activity for robots interacting with dynamic elements in their environment, for instance inspection of pieces moving along a 
conveyor belt or pursuit of other robots. Object tracking has also been employed in other fields, for instance tracking moving bacteria seen through a 
microscope~\citep{wood2012}.

Over the years, the robotics community has produced various object trackers with properties catering to different circumstances. For example, pedestrian 
tracking~\citep{choi2011detecting},~\citep{wu2005statistical} using RGB or RGB-D data for surveillance purposes. Or, for vehicle tracking via RGB data or 2D/3D 
range data for counting number of vehicles on the road and for traffic congestion control. These kinds of object-specific trackers are generally trained 
offline given that the shape and/or motion model of the object of interest is known in advance. Our work falls in between object-specific trackers and generic 
object trackers. That is, our object tracker assumes that the appearance model of the object of interest is known in advance, but the motion model is not known. 
It should also be noted that even in the absence of an appearance model of the object, our proposed method still tracks the object but not accurately. The 
remainder of the paper is described assuming that the appearance model of the object of interest is known in advance.

Our work addresses the problem of tracking fast moving targets in mobile robotics. These targets can be anything from Unmanned Aerial Vehicles to ground 
rovers. As these targets have complex motion, it is not feasible to assume a fixed motion model in advance. Moreover, due to various effects such as occlusion 
or motion blur, track of the object might be lost and hence generic object trackers~\citep{GFF18}, \citep{bertinetto2016fully}, \citep{held2016learning}, 
\citep{valmadre2017end} cannot be used as they cannot re-identify objects once track is lost. Therefore, we propose an object tracking system which can track 
fast moving objects and also has the ability to regain track of the object if required.

To tackle the problem of visual object tracking we take inspiration from evolution, which has enabled humans and other animals to survive in their ecosystem 
and effectively interact with their environment. The Human visual system outperforms artificial visual systems primarily because of the brain's ability to 
perform global data association. Moreover, performing robust target tracking is a complex task which requires multiple subsystems working together. Thus, our 
target tracking system is composed of individual subsystems for solving different problems that arise in tracking of fast moving targets. 

In this paper, we present a tracker that uses a fusion of motion tracking and object detection. Our algorithm is governed by a 
heuristic but its major sub-components are adaptive learning models. The main contributions of the work are:
\begin{enumerate}
 \item Formulation for target tracking based on a heuristic fusion of a generic object tracker~\citep{GFF18} with a detector~\citep{redmon2017yolo9000}.
 \item Reducing the impact of motion blur while tracking fast moving objects.
\item Demonstrating that our approach to object tracking significantly improves the quality of tracking with respect to the performance obtained solely by the 
motion tracker and object detector modules used in our algorithm. 
 \end{enumerate}
We provide the full implementation of our algorithm, whose source code is written in Python. We provide two versions of this code: one as a standalone Python 
program, and another as a ROS package ready for use in robotic applications. The code is available at \url{https://bitbucket.org/Pranoy97/fusion}.

%\section{RELATED WORK}

%Object tracking is one of the most important tasks in a majority of activities in mobile robotics where we need to follow a master to mimic its activities or 
%to follow a target for pursuit. In the past, the tracking algorithms were mainly based on predefined motion or shape models of the object or, a particular 
%heuristic was to followed to track moving objects.
%***** describe about the heuristic based algorithms******

%****** in next para, explain the learning based algorithms ****

%****** in the next para, explain why for real world application,
%only learning based methods cannot be used and hence briefly state the importance and uniqueness of our algorithm *******

\section{Method}
Our object tracking pipeline consists of three modules: an Object and Blur Detection module, a Motion Tracking module and a Re-identification module. The 
proposed method takes an initial bounding box as input from the user and then starts the tracking procedure. The object enclosed in the bounding box is 
classified as either a \emph{familiar} (seen before by the object detection module) or \emph{unfamiliar} object. Our work deals only with the \emph{familiar} 
object case, but for the sake of completeness we handle the \emph{unfamiliar} case by deploying only the motion tracking algorithm.

We hypothesize that for tracking objects that follow complex trajectories, better tracking can be achieved by interim tracker initialization i.e.~the tracker 
is initialized after every $n$ frames. The intuition behind this hypothesis is that, if $n$ is sufficiently small, then the tracker has to essentially learn a 
piece-wise linear model of the motion. Thus, this initialization procedure greatly reduces the location estimation error and drift error of the motion tracker.

In the case of a \emph{familiar} object, in accordance with our hypothesis, our object detection module is run on every $n^\text{th}$ frame while the tracking 
algorithm is employed in the $n-1$ frames in between. Validation of the track of the object of interest is done by the object detection module. If the track is 
lost, the object detection module scans the entire frame and reports all the objects in the scene that belong to the class of the object of interest. For 
example, if the object of interest is a human, then all the humans in the scenes are reported. Then, we invoke the re-identification module to find the best 
possible match. If the re-identification score for the best match is above an empirically determined threshold, we declare that the object of interest is 
re-identified. Else, we keep on scanning the frame until we find the object. When the object of interest is re-identified, we resume the normal tracking 
procedure described previously. 

The remainder of this section describes the exact formulation of each of the modules and the pseudo-code for our algorithm.

\subsection{Object and Blur Detection module}

Existing object tracking algorithms inherently track the motion of an individual target. This means they will lose track of the object in the event of overly 
long occlusion or if the object moves faster than the tracking algorithm can handle. It is also common for existing tracking algorithms to accumulate errors 
such that the bounding box tracking the object slowly drifts away from the object it is tracking. To fix these problems we propose to do two things. First, to 
add an Object Detector. Here, we run a fast and accurate object detector (we use YOLO v2~\citep{redmon2017yolo9000} in our work) on every $n^\text{th}$ frame 
while the motion tracking module is employed in the $n-1$ frames in between. This means that the tracking procedure is initialized in every $n^\text{th}$ 
frame. The value of $n$ is chosen such that it is small enough to prevent the accumulation of error and at the same time large enough to enable the tracker to 
adapt to the object’s motion.

Secondly, we introduce a Blur detector. Since motion blur significantly hinders the tracking of fast moving objects, we quantitatively evaluate a metric which 
gives the measure of the amount of motion blur in the neighbourhood of the object of interest. Specifically, let $X_c,Y_c$ be the center of the bounding box of 
the object of interest at time $t$, and $W,H$ be respectively the width and height of the bounding box. Then, the region of interest (ROI) is defined as the 
crop of the input image at time $t+1$ centered at $X_c,Y_c$ and having width and height as $2W,2H$ respectively. 

A blurry image has fewer sharp edges than a less blurry version. This implies that in the frequency domain, high frequency components (which contribute to 
sharp edges) have a lower magnitude for blurry images. To evaluate the high spatial frequencies associated with sharp edges in an image, we use the variance of 
Laplacian method~\citep{pech2000diatom}. It quantitatively evaluates the amount of blur in a given frame, and if the blur is found to be greater than a 
threshold value then we deem that frame to be blurry. The exact formulation of the method is given next.

The Laplacian mask $l$ is defined as
\begin{equation}
l
=
\frac{1}{6}
\begin{pmatrix} 
0 & -1 & 0 \\ 
-1 & 4 & -1 \\ 
0 & -1 & 0  
\end{pmatrix} 
\end{equation}
Now, let $L$ be the  convolution of  the  input  image $I$ with the mask $l$, and $(M,N)$ be the resolution of image $I$. The variance of Laplacian is defined 
as follows
\begin{equation}
LAP\_{VAR}(I) = \sum_{m=1}^{M}\sum_{n=1}^{N}\big[|L(m,n)|-\bar{L}\big]^{2}
\end{equation}

where $\bar{L}$ is the mean of absolute values given by
\begin{equation}
\bar{L} = \frac{1}{MN}\sum_{m=1}^{M}\sum_{n=1}^{N}|L(m,n)|
\end{equation}

The entire algorithm to check whether a frame is blurry or not is as follows:

\begin{algorithm}[H]
% the below written 2lines does the following
%The first one \renewcommand{\thealgorithm}{} removes the numbering, %while the second one \floatname{algorithm}{} removes the word %'Algorithm'
\renewcommand{\thealgorithm}{}
\floatname{algorithm}{}
\caption{\textbf{IsImageBlurry} function}
\scriptsize{
\begin{algorithmic}[1] % here the [1] enables the numbering of the lines in the algorithm
\State{$(X_c,Y_c) \gets $} Center of bounding box
\State{$(W,H) \gets$ } Width and Height of bounding box
\State {$ROI \gets$} $I \big[ (X_c - W) : (X_c + W), (Y_c - H) : (Y_c + H) \big]$
\If{$LAP\_VAR(ROI) > blurThresh$}
    \State {\textbf{return} True}   
\Else
    \State{\textbf{return} False}
\EndIf
\end{algorithmic}
}
\end{algorithm}

In summary, we propose that when a blurry image is detected(via the \textbf{IsImageBlurry} procedure), the object detection algorithm (which is a priori 
trained on both blurry and non-blurry images) is employed on that particular frame and the tracker module is re-initialized. Because of its training, the 
object detection algorithm can detect blurry objects up to a certain extent and therefore help improve the tracking performance of fast-moving objects.

\subsection{Motion tracker module}

For learning the motion model of the object of interest we rely on $\text{Re}^3$~\citep{GFF18} which was generously open-sourced by Gordon \emph{et. al}. The 
$\text{Re}^3$ algorithm was trained for generic object tracking. Its recurrent model gives it the ability to train from motion examples offline and quickly 
update its recurrent parameters online when tracking specific objects. It is also very fast hence well suited for implementation on single-board computers 
commonly used in mobile robotics applications.

$\text{Re}^3$ employs a Recurrent Neural Network composed of Long Short-term memory (LSTM) units. RNN’s are well suited for sequential data modeling and 
feature extraction~\citep{gravessupervised}. The recurrent structure and the internal memory of RNN facilitate its modeling of the long-term temporal dynamics 
of sequential data. LSTM is an advanced RNN architecture which mitigates the vanishing gradient effect of RNN. The basic formulation of an LSTM unit is given 
in Equations~\eqref{eq:LSTM} where $t$ represents the frame index, $x_t$ is the current input vector, $h_{t-1}$ is the previous output(or recurrent) vector 
$W$, $U$ are weight matrices for the input and recurrent vector respectively, $b$ is the bias vector, $\sigma$ is the sigmoid function, and $\circ$ is 
point-wise multiplication. A forward pass produces both an output vector $h_t$, which is used to regress the current output, and the cell state $c_t$, which 
holds important memory information. Both $h_t$ and $c_t$ are fed into the following forward pass, allowing for information to propagate forward in time.
%equations for LSTM
\begin{equation}
\begin{aligned}
f_t = \sigma_g(W_fx_t+U_fh_{t-1}+b_f) \\
i_t = \sigma_g(W_ix_t+U_ih_{t-1}+b_i) \\
o_t = \sigma_g(W_ox_t+U_oh_{t-1}+b_o) \\
c_t = f_t \circ c_{t-1} + i_t \circ \sigma_c(W_cx_t+U_ch_{t-1}+b_c) \\
h_t = o_t \circ \sigma_h(c_t)
\end{aligned}
\label{eq:LSTM}
\end{equation}

For the purpose of motion tracking, the recurrent parameters in the LSTM network represent the tracker state which can be updated with a single forward pass. 
In this way, the tracker learns to use new observations to update the motion model, but without any additional computation spent on online training. RNN's 
accept features as inputs and extract useful temporal information from those. But, they are not good at extracting useful information from raw images or video 
sequences. Hence, $\text{Re}^3$ uses a Convolutional Recurrent Neural Network (CRNN) for the task of motion tracking.

\subsection{Re-identification module}

The Object tracker can lose track of the target due to various reasons such as long term occlusions, fast motions, or other factors. To achieve robust 
tracking, it is necessary to have a system which can re-identify the object of interest. This system needs to be able to recognize objects in real-time. But, as 
the object is not known in advance, image recognition algorithms cannot be used to complete the task. Moreover, the entire pipeline for tracking has to be very 
fast, which means computationally intensive algorithms cannot be employed for object re-identification. Therefore, we propose a fast and fairly accurate 
procedure, which uses color cues and structural similarity to re-identify the object of interest.
Here, we assume that the track of the object is not lost in the very first second. So, if the frame rate of the camera recording the video is $k$, then we have 
$k$ bounding boxes of the object of interest. We store these $k$ bounding boxes and henceforth refer to them as templates. 
Now, to perform the re-identification task we evaluate two scores; one from normalized cross-correlation~\citep{lewis1995fast} of the frame with the stored 
templates, and the other from color histogram matching between the templates and
the current frame. Then we take a sum of these two scores to evaluate the re-identification score. The exact formulation is explained below:

\textbf{Normalized Cross Correlation}:
Let $f$ be the current frame, and define a sum over $x, y$ within the window containing the template $t$ positioned at point $(u, v)$. $\bar{t}$ is the mean of 
$t$ and $\bar{f}(u,v)$ is the mean of $f(x,y)$ in the region under the template $t$. We calculate $\gamma(u,v)$, the correlation coefficient which gives the 
measure of the normalized correlation between the current frame and the template:

% equation for Normalized Cross. Corr.
\begin{equation}
    \begin{aligned}
    &\gamma(u,v) \\
    &= \frac{\sum_{x,y}[f(x,y)-\bar{f}_{u,v}][t(x-u,y-v),\bar{t}]}{[\sum_{x,y}[f(x,y)-\bar{f}_{u,v}]^2\sum_{x,y}[t(x-u,y-v),\bar{t}]^2]^{0.5}}
    \end{aligned}
\end{equation}

The correlation coefficient is a measure of similarity between the frame and the template. It is invariant to the template size and changes in amplitude such 
as those caused by changing lighting conditions across the image sequence. Due to the above characteristics, this metric of evaluation of structural similarity 
is robust to noise and brightness change in the scene and hence suited for re-identifying objects in noisy environments.

The templates contain different view-points of the object of interest. So, in order to check if the object under consideration is actually the object of 
interest, we take the max of correlation coefficients evaluated between the current frame and the templates. We refer to this score as \emph{NCC\_score}.

\textbf{Colour Histogram Intersection Algorithm} \citep{swain1992indexing}:
Colour helps the human visual system to analyze complex images more efficiently, improving object recognition. Psychological experiments have shown that color 
gives a strong contribution to memorization and recognition~\citep{wichmann2002contributions} in humans. Hence, to leverage from the above observations made by 
researchers in the field of cognitive science, we use the color histogram intersection algorithm as a metric to re-identify the object of interest in the scene.
The histogram intersection algorithm was proposed in \citep{swain1991color}. This algorithm does not require the accurate separation of the object from its 
background and it is robust to occluding objects in the foreground. Also, histograms are invariant to translation and they change slowly under different view 
angles, scales and in presence of occlusions, hence they are well suited for our purpose of object re-identification. 

The mathematical formulation of the algorithm is as follows: given the histogram $i$ of the input image $I$ and the histogram $m$ of a template $t$, each one 
containing $n$ bins, the intersection $Z$ is defined as:

% equation
\begin{equation}
    \begin{gathered}
        Z = \sum_{j=1}^n \text{min} (i_j,m_j)
    \end{gathered}
\end{equation}

The result of the intersection is the number of pixels from the template that have corresponding pixels of the same colors in the input image. To normalize the 
result between 0 and 1 we have to divide it by the number of pixels in the template histogram:

% equation
\begin{equation}
    \begin{gathered}
       \hat{Z} = \frac{\sum_{j=1}^n \text{min}(i_j,m_j)}{\sum_{j=1}^n m_j}
    \end{gathered}
\end{equation}

When an unknown object image is given as input the intersection algorithm computes the histogram intersection for all the stored templates.

The complete re-identification module can be summarized as follows:
\begin{algorithm}[H]
% the below written 2lines does the following
%The first one \renewcommand{\thealgorithm}{} removes the numbering, %while the second one \floatname{algorithm}{} removes the word %'Algorithm'
\renewcommand{\thealgorithm}{}
\floatname{algorithm}{}
\caption{\textbf{Re-identification} function}
\scriptsize{
\begin{algorithmic}[1] % here the [1] enables the numbering of the lines in the algorithm
\State{$reIdentification\_score \gets $} $\hat{Z}$ + \emph{NCC\_score}
\If{$reIdentification\_score > \epsilon$}
    \State { $trackFound \gets True$}   
\EndIf
\end{algorithmic}
}
\end{algorithm}
where $\epsilon$ is a constant associated with the number of templates used. $\epsilon$ = 1.2 is found experimentally.
\subsection{Fusion Algorithm pseudo-code}

Our overall Fusion algorithm can be summarized as follows:

\begin{algorithm}[H]
\renewcommand{\thealgorithm}{}
\floatname{algorithm}{}
\caption{\textbf{Fusion} algorithm}
\scriptsize{
\begin{algorithmic}[1] % here the [1] enables the numbering of the lines in the algorithm
\State Start
\State Initialize the motion tracker with $bbox$.
\State{$N \gets 0$} \Comment{Counter for number of frames}
\State Run the object detector on the $ROI$, set the $known$ flag
\State{$LostTrack \gets False$} \Comment{ Variable for keeping Track of object }
\While{$ True $}
\State{$ N \mathrel{{+}{=}} 1$}
\If{$known$}
    \If{$LostTrack$}
        \State{$bboxes \gets $} ObjectDetector(Image) \Comment{Scan for object of interest}
        \If{len$(bboxes) \geq 0$}
        % \State{$bbox \gets$} bounding box with max re-identification score
        \State{$bbox \gets$ Re-identification(bboxes)} \Comment{re-identification}
        \If{$bbox$ \textbf{is not} $None$}
            \State{$LostTrack \gets False$} \Comment{ Object re-identified}
        \EndIf
        \Else
            \State{$bbox \gets $} MotionTracker.track(Image) \Comment{Motion Tracker}
        \EndIf
    \Else
        \If{(mod(N,Threshold) = 0) $\mid $ IsImageBlurry } \Comment{Blur Detector}
            \State{$bbox \gets $} ObjectDetector(Image) \Comment{Object Detector}
            \If{$bbox$ \textbf{is} $None$}
                \State{$LostTrack \gets True$} \Comment{ Object track lost}
            \EndIf
        \Else
            \State{$bbox \gets $} MotionTracker.track(Image) \Comment{Motion Tracker} 
        \EndIf  
    \EndIf    
\Else
    \State{$bbox \gets $} MotionTracker.track(Image) \Comment{Motion Tracker}
\EndIf
\EndWhile
\end{algorithmic}
}
\end{algorithm}

\section{Experimental Results}

We compare our fusion algorithm to other tracking methods on two popular tracking datasets in terms of overall performance.

As our work focuses on improving the performance in object tracking for the specific case of the object of interest being \emph{familiar}, many of the video 
sequences on which we compare our algorithm with other existing algorithms contains \emph{familiar} objects. We demonstrate our algorithm's effectiveness by 
testing on standard tracking benchmarks, the Visual Object Tracking 2017 challenge~\citep{kristan2017visual} and Online Tracking Benchmark 
(OTB)~\citep{wu2013online} datasets. 

\subsection{Methodology for quantitative evaluation}

We use the One Pass Evaluation (OPE) criterion for evaluating performance on both of the benchmark datasets. This is the conventional way to evaluate trackers 
which are run throughout a test sequence with initialization from the ground truth position in the first frame, then reporting the average precision or success 
rate. This is referred as one-pass evaluation (OPE)\citep{wu2013online}. It consists of two separate plots:

\begin{enumerate}
\item \textbf{Precision plot}: This shows the percentage of frames whose estimated location is within the given threshold distance of the ground truth. As the 
representative precision score for each tracker we use the score for the threshold = 20 pixels.
\item \textbf{Success Plot}: The second metric is the bounding box overlap. Given the tracked bounding box and the ground-truth bounding box, we evaluate the 
Intersection over Union (IoU). To measure the performance on a sequence of frames, we count the number of successful frames whose overlap $S$ is larger than 
the given threshold (here, the threshold was kept $0.5$ or $50$ percent overlap). The success plot shows the ratios of successful frames at the thresholds 
varied from $0$ to $1$.
\end{enumerate}

\subsection{VOT 2017\citep{kristan2017visual} and OTB-50\citep{wu2013online}}

The VOT 2017 object tracking dataset consists of 60 videos, made explicitly for the purpose of testing object trackers. We use 24 of those video sequences, 
which have most of their objects belonging to the \emph{familiar} class, with some from the \emph{unfamiliar} class. Many of the videos from this dataset 
contain difficulties such as large appearance change, heavy occlusion and camera motion. For fairness of comparison, all testing and benchmarking was performed 
on the same computer.

\begin{figure}[hpbt!]
    \centering
     \includegraphics[width=\linewidth]{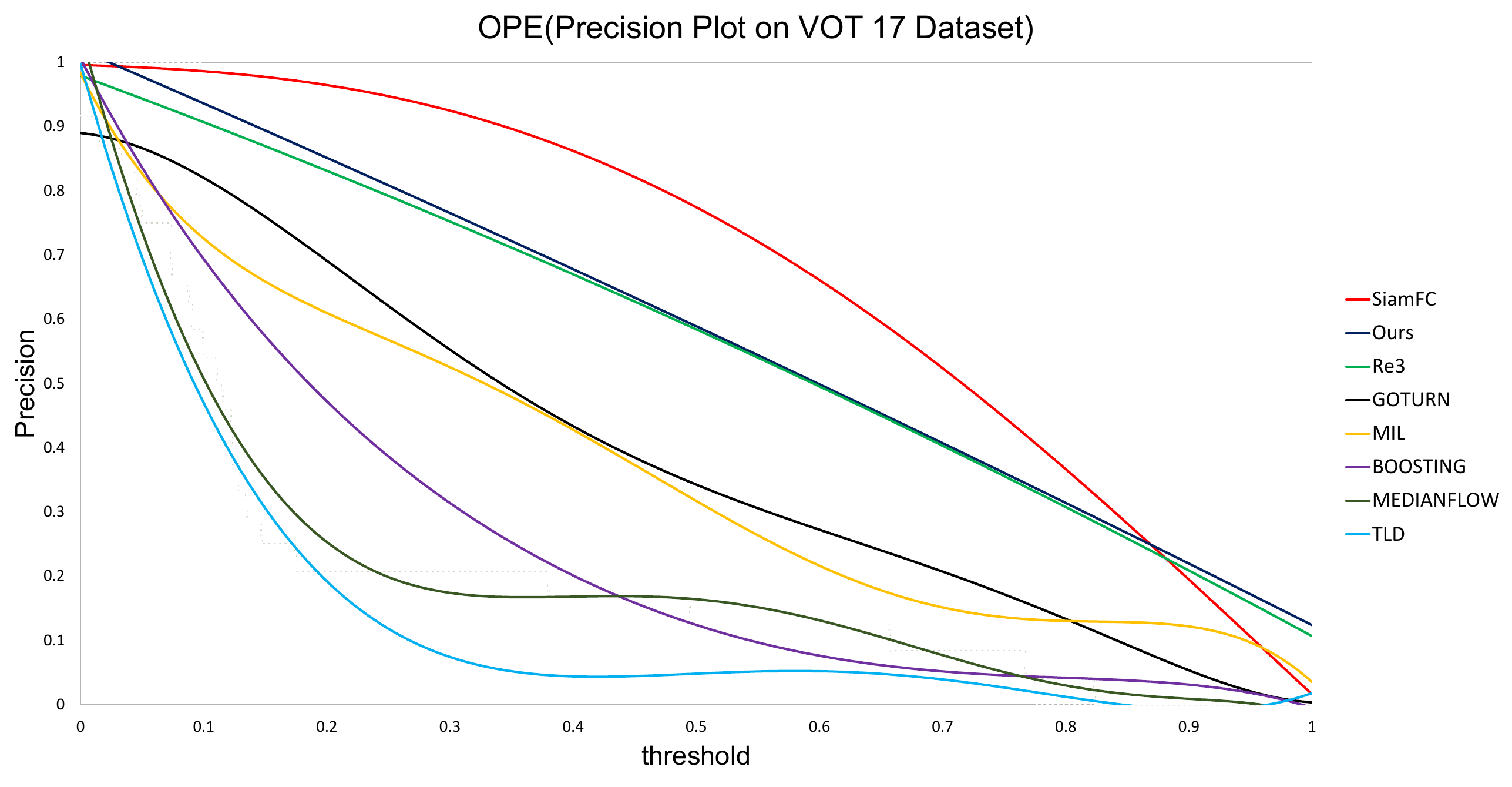}
  \caption{Precision Plot for VOT17}
 \label{fig:pplotVOT}
\end{figure}

\begin{figure}[hpbt!]
    \centering
     \includegraphics[width=\linewidth]{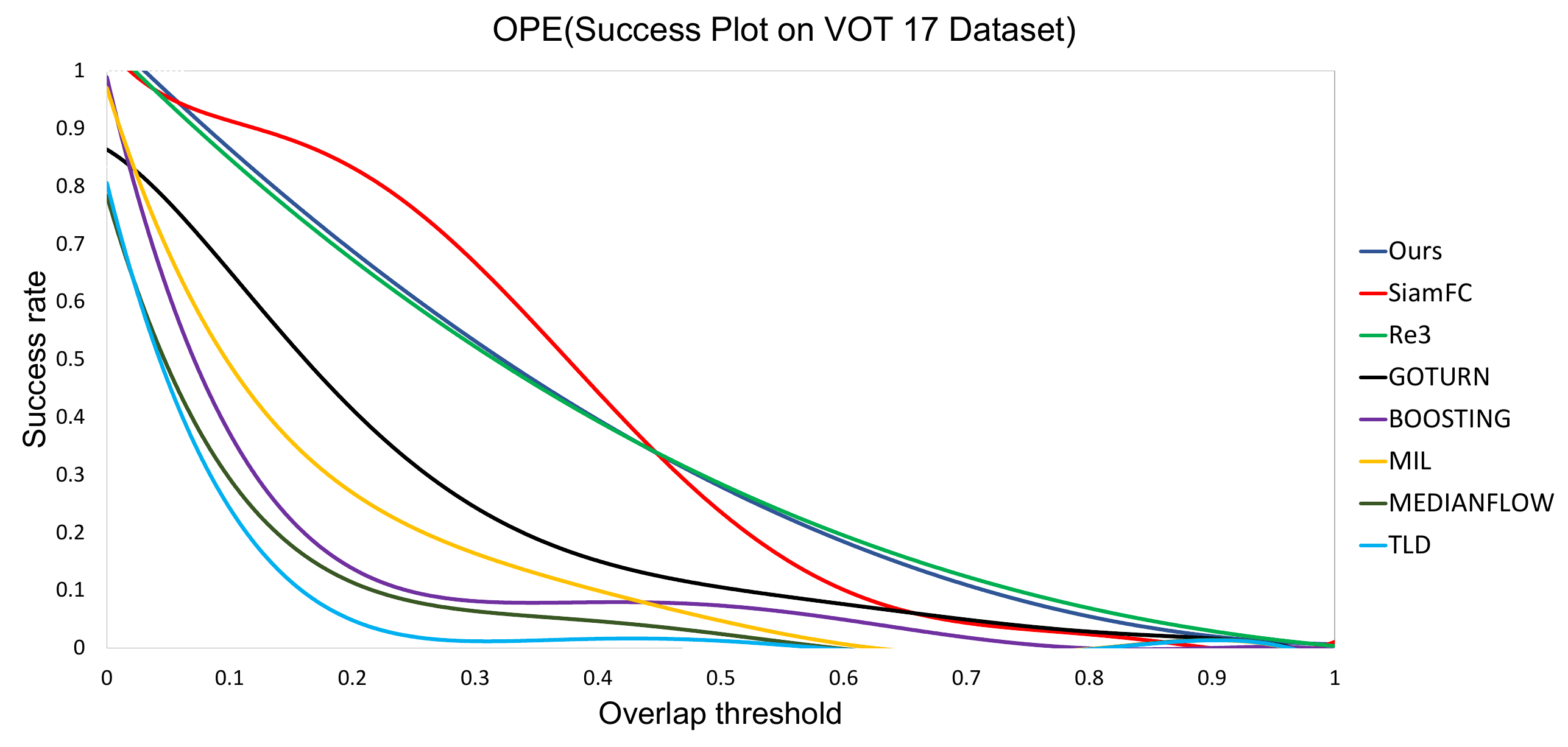}
  \caption{Success Plot for VOT17}
 \label{fig:splotVOT}
\end{figure}

\begin{figure}[hpbt!]
    \center
     \includegraphics[width=\linewidth]{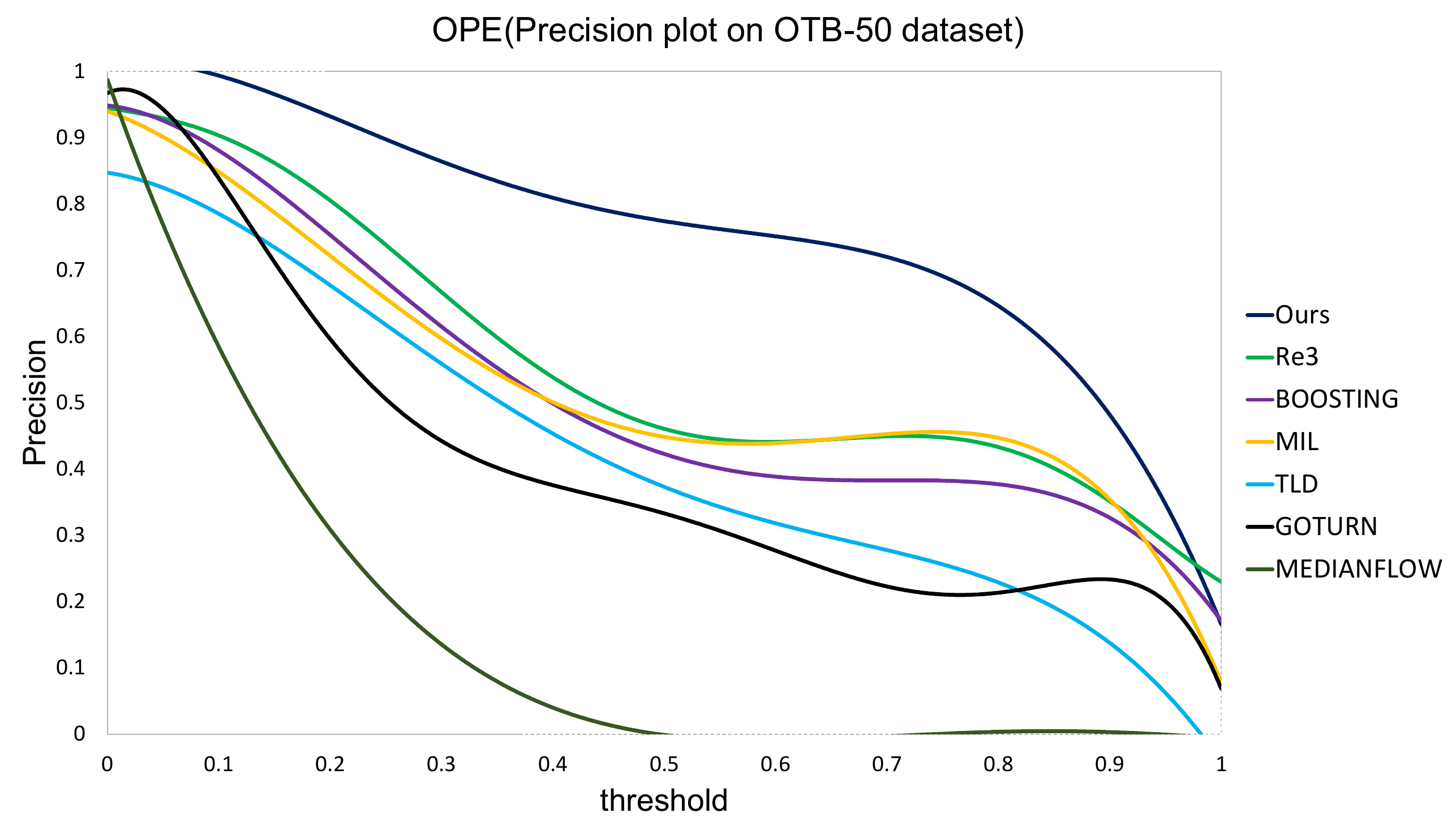}
  \caption{Precision Plot for OTB-50}
 \label{fig:pplotTB}
\end{figure}

\begin{figure}[hpbt!]
    \center
     \includegraphics[width=\linewidth]{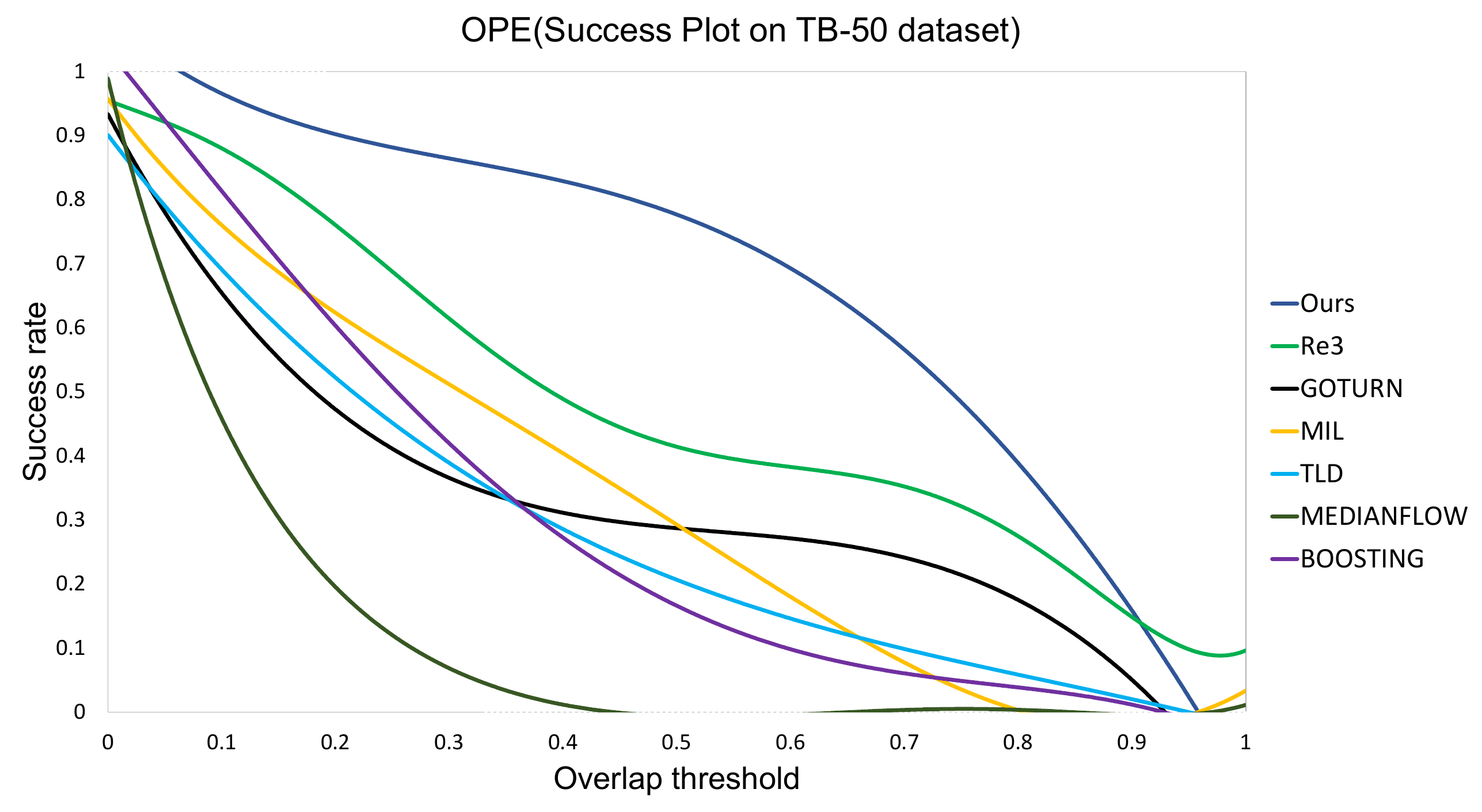}
  \caption{Success Plot for OTB-50}
 \label{fig:splotTB}
\end{figure}

Figure~\ref{fig:pplotVOT} compares our proposed method with other trackers including CFNet~\citep{valmadre2017end}, the winner of the VOT17 real time 
challenge~\citep{valmadre2017end}. We also tested our method on a set of 9 videos (all containing objects from the \emph{familiar} class) from the OTB dataset. 
Figures~\ref{fig:pplotVOT}--\ref{fig:splotTB} compare our proposed method with other trackers. Note the performance improvement in both the Success Rate and 
Precision of our method over the $\text{Re}^3$ method by itself is significant. The corresponding numerical results are given in 
Table~\ref{tab:compresults}.

%set of figures for evaluation

\begin{table}[htbp!]
\centering
\caption{Comparison Table (overlap threshold = 0.5, precision threshold = 0.5)}
\label{tab:compresults}
\begin{tabular}{|c|c|c|c|}
\hline
Name & Success Rate & Precision & FPS \\
\hline
Ours & $\textbf{0.3333}$ & $0.5833$ &$86$\\
SiamFC & $0.2083$ & $\textbf{0.7916}$ &$45$\\
Re3    & $\textbf{0.3333}$ & $0.5833$ &$120$\\
GOTURN & $0.125$  & $0.375$  &$60$\\
BOOSTING & $0.0833$ & $0.1666$ &$66$\\
MIL    & $0.0833$ & $0.333$  &$25$\\
MEDIANFLOW & $0.0001$ & $0.125$ &$\textbf{250}$\\
TLD    & $0.0001$ & $0.0416$ &$30$\\
\hline
\end{tabular}

\end{table}

\subsection{Qualitative Results}

\begin{figure}[hpbt!]
    \centering
    \begin{tabular}{c c c c c}
     \includegraphics[width=0.15\linewidth]{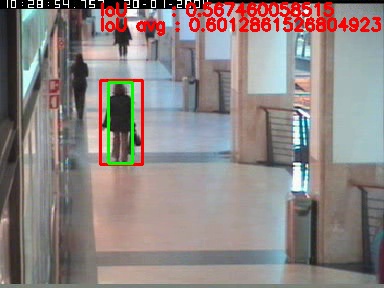}
 &     \includegraphics[width=0.15\linewidth]{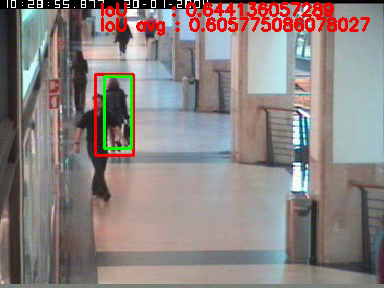}
 &      \includegraphics[width=0.15\linewidth]{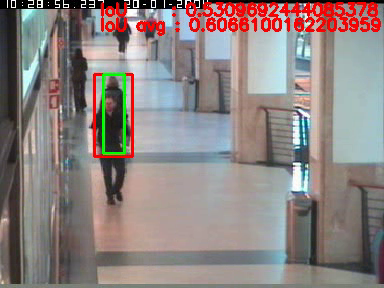}
 &      \includegraphics[width=0.15\linewidth]{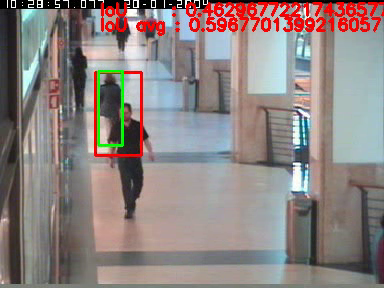}
 &     \includegraphics[width=0.15\linewidth]{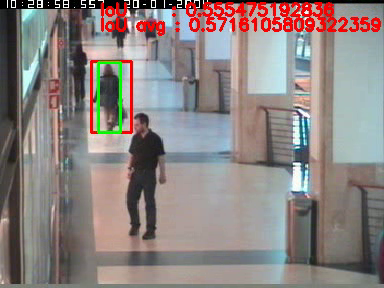} \\
 (a) t=1 & (b) t=2 & (c) t=3 & (d) t=4 & (e) t=5 
\end{tabular}
\caption{Tracking of occluded pedestrian}
\label{fig:res_flag}
\end{figure}

\begin{figure}[hpbt!]
    \centering
    \begin{tabular}{c c c c c}
     \includegraphics[width=0.15\linewidth]{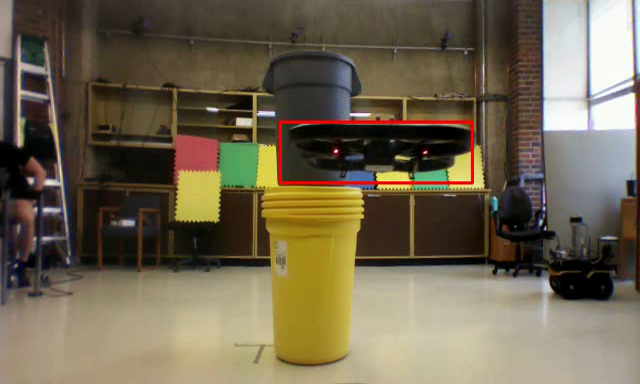} 
 &      \includegraphics[width=0.15\linewidth]{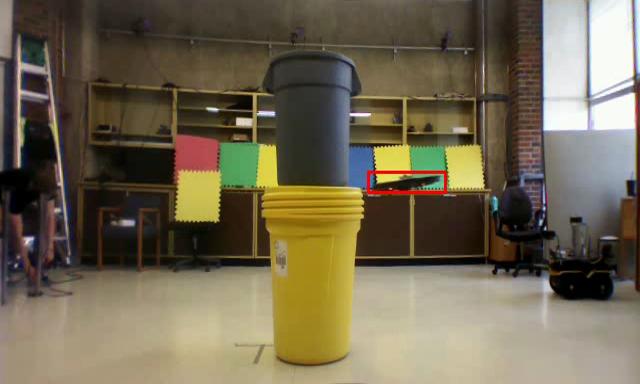}
 &      \includegraphics[width=0.15\linewidth]{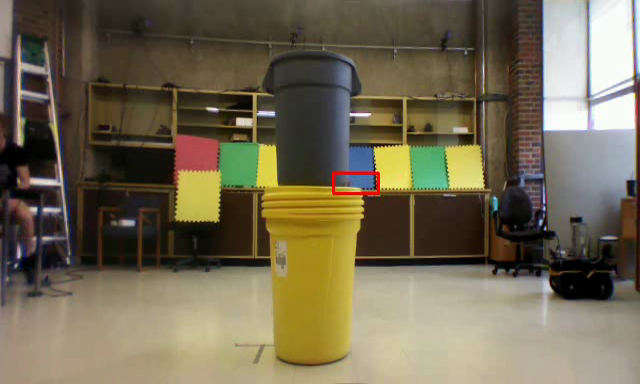}
 &      \includegraphics[width=0.15\linewidth]{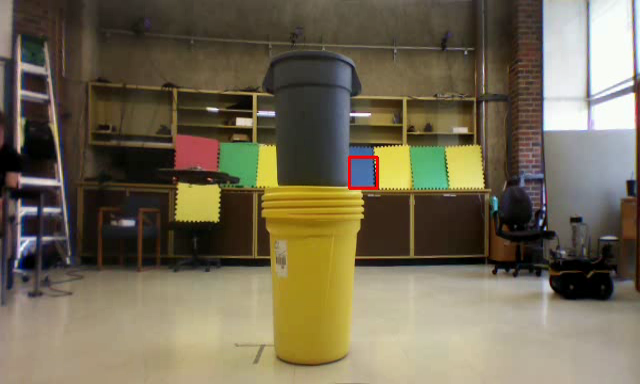}
 &     \includegraphics[width=0.15\linewidth]{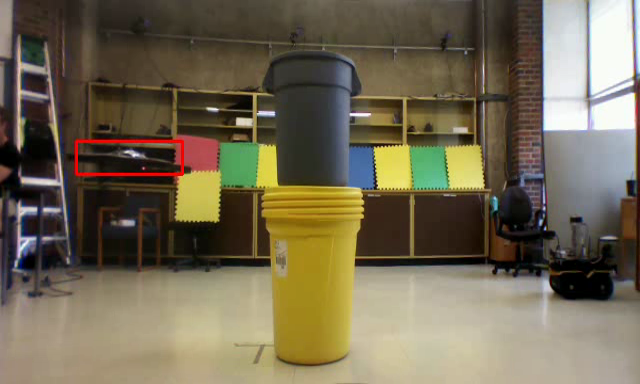} \\
 (a) t=1 & (b) t=2 & (c) t=3 & (d) t=4 & (e) t=5 
\end{tabular}
    \caption{Tracking of flying UAV drone}
    \label{fig:res_flag2}
\end{figure}

We also tested our algorithm on a variety of challenging videos in order to ascertain its usefulness in the real world. Our method performs remarkably well on 
challenging tasks such as tracking a fast moving drone, tracking people performing acrobatics and stunts, or simultaneously tracking multiple people ($\leq 
4$). 
We show excerpts from two such videos in Figures~\ref{fig:res_flag} and~\ref{fig:res_flag2}.

\section{Conclusion}

In this paper, we presented a learning-agnostic heuristic for robust object tracking. Our work fuses the previous works of $\text{Re}^3$~\citep{GFF18} and 
\emph{YOLO v2}~\citep{redmon2017yolo9000} to build a robust vision-based target tracker usable for mobile robotic applications. Both modules are lightweight 
and capable of real-time performance on portable computing platforms typical in mobile robotics. Our fusion method demonstrates increased accuracy, robustness 
and speed compared to other trackers, especially during periods of occlusion or high speed motions of the target or the camera. We demonstrated that provided 
the target belongs to the \emph{familiar} category of objects, our proposed algorithm provides accuracy similar or better than the current state of the art 
trackers while being able to operate at an average at 85 FPS, faster than most trackers. 

% \begin{ack}
% Place acknowledgments here.
% \end{ack}

\bibliography{bibliography}             % bib file to produce the bibliography
                                                     % with bibtex (preferred)

\end{document}